%% file: main.tex
\documentclass{article}
\usepackage{spconf,amsmath,graphicx}

\usepackage{algorithm}
\usepackage{algorithmic}
\usepackage{amssymb}
\usepackage{url}

\title{Deep Temporal Interpolation of Radar-based  Precipitation}
\name{\begin{tabular}{c}Michiaki Tatsubori \qquad
	Takao Moriyama \qquad
	Tatsuya Ishikawa \qquad
	Paolo Fraccaro \\
	Anne Jones \qquad
	Blair Edwards \qquad
	Julian Kuehnert \qquad
	Sekou L. Remy
\end{tabular}}
\address{IBM Research}

\begin{document}
\maketitle
\begin{abstract}
When providing the boundary conditions for hydrological flood models and estimating the associated risk, interpolating precipitation at very high temporal resolutions (e.g. 5 minutes) is essential not to miss the cause of flooding in local regions.   In this paper, we study optical flow-based interpolation of globally available weather radar images from satellites.  The proposed approach uses deep neural networks for the interpolation of multiple video frames, while terrain information is combined with temporarily coarse-grained precipitation radar observation as inputs for self-supervised training.  An experiment with the Meteonet radar precipitation dataset for the flood risk simulation in Aude, a department in Southern France (2018), demonstrated the advantage of the proposed method over a linear interpolation baseline, with up to 20\% error reduction.
\end{abstract}
\section{Introduction}

% Motivating the satellite observations
Satellite observations which can consistently capture broad areas~\cite{Skofronick-Jackson+:2017:GPM} are unique and effective means to achieve global scale rainfall measurement~\cite{deWitt+:AAAI2021:RainBench}.
While ground-based rainfall observation networks using rain gauges~\cite{Funk+:2015:CHIRPS} or weather radars provide dense and highly frequent observation such as every 10 minutes, ground observation data is often missing in Asia and African regions~\cite{Bai+:2018:CHIRPS_China,Tufa+:2018:CHIRPS_Africa} even though these regions suffer from many water related disasters.

%Satellite-based precipitation data would be the only rainfall information in severe environments, where humans cannot enter, and in troubled regions. In addition, international rivers running through a number of countries are one of a source of conflicts regarding water acquisition, since countries located in the lower reaches of international rivers have serious problems in obtaining precipitation data from countries in the upper reaches.

% https://www.eorc.jaxa.jp/GPM/en/overview.html

% Impact of precipitation use
Precipitation is one of the main causes of fatal natural disasters, as it causes flooding, landslides, and snow avalanches as well as damage to crops. The frequency and severity of these events is increasing with the increase in extreme weather events associated with acute climate change. Mathematical and computational models such as IFM~\cite{Singhal+:EuroPar2014:IFM} have become widely used tools to predict and mitigate risks to socioeconomic systems. However, for these predictions to be accurate, the underlying precipitation data driving the simulations have to be well resolved in space and time. 
Forecasting precipitation at very high resolutions in space and time, particularly with the aim of providing the boundary conditions for hydrological models and estimating the associated risk, is one of the most difficult challenges in meteorology~\cite{Marrocu+Luca:Forecasting:2020}.
%In short, the problem is to predict the field of precipitation on grids of 1 km or less and for horizons of less than a few hours. Due to the high spatial resolutions required, methods based on meteorological models are not effective, because they are too onerous computationally and the time to perform a simulation is usually excessively too long for operational purposes. They also depend, in a critical way, on the initial conditions as the level of uncertainty is too high and it is difficult to assimilate locally recorded data.
%Even if these difficulties were overcome, a fundamental difficulty would remain, the gap between the spatial scales provided by the meteorological models and those useful to hydrologists for operational purposes. To overcome this difficulty and quantify the uncertainty involved, probabilistic approaches were proposed.

\begin{figure}
	\centering
	\includegraphics[width=1.0\columnwidth]{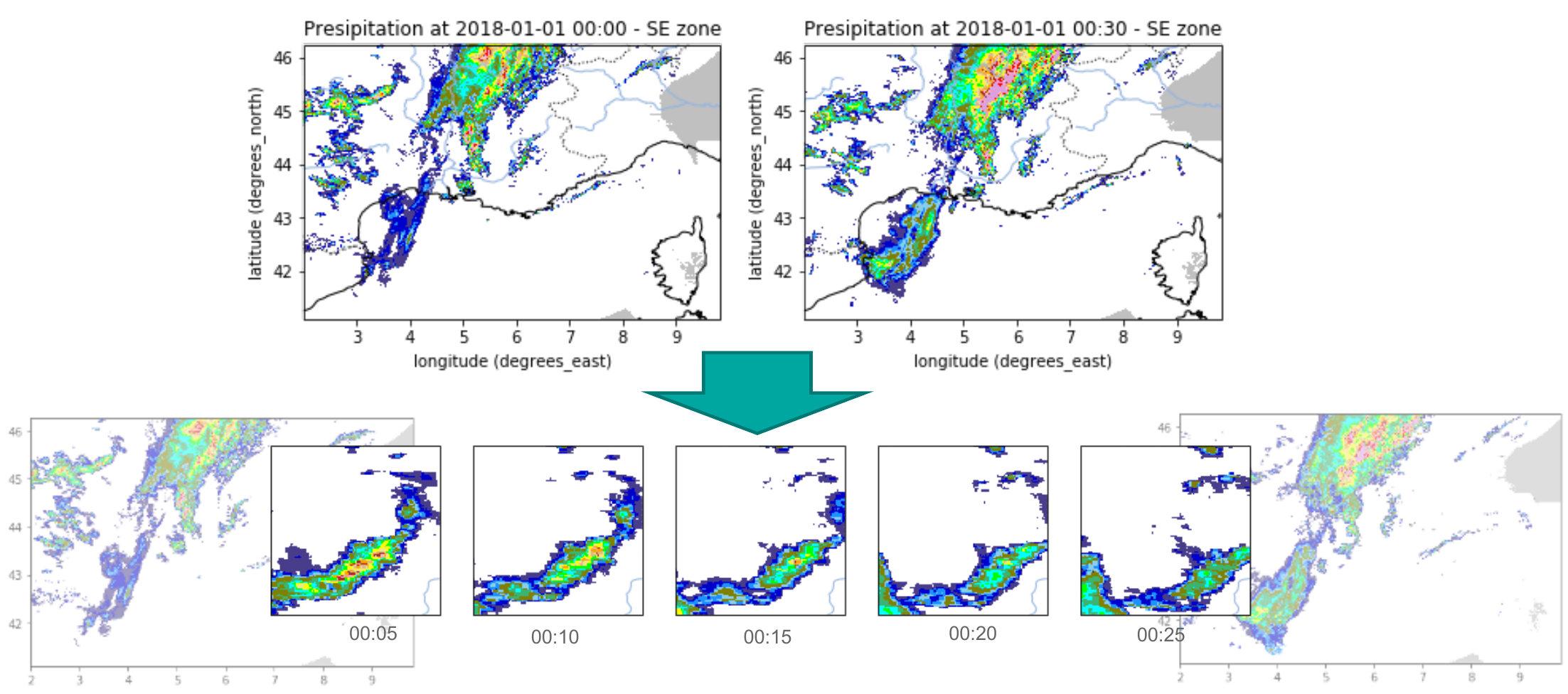}
	\caption{An example task of temporal interpolation of precipitation.  Given half-hourly snapshots from radar observation, we produce precipitation snapshots of 5 minutes.}
	%\caption{Example task of temporal interpolation of precipitation.  Given half-hourly snapshot from radar observation, we produce precipitation snapshots of 5 minutes, which can be used in a further pipeline of climate impact modeling, such as flooding.}
	\label{fig:teaser}
\end{figure}

This paper introduces a temporal precipitation interpolation method based on deep learning and demonstrates how the temporal resolution of it impacts an actual flood modeling results.
Along with the recent work with deep learning for nowcasting~\cite{Bechini+Chandrasekar:2017:Nowcasting} with topographic feature learning~\cite{Proulx-Bourque+:IGARSS2018:topographic_feature}, our focus is on the temporal interpolation of precipitation for climate impact modeling.  The task definition is illustrated as Figure~\ref{fig:teaser}.  The results would be used to calibrate the impact modeling to be accurate in forecasting the disaster risks in the future, which is the motivation of the work but out of the scope for this paper.

%% A "teaser" image appears between the author and affiliation
%% information and the body of the document, and typically spans the
%% page.

% contribution of the work

%Research question we try to answer in this paper is as follows.
%%
%\begin{itemize}
%\item Can we leverage the existing deep video interpolation technology for temporal precipitation interpolation, to have better results than simple linear interpolation, which is usually used in climate impact modeling?
%\item Can we leverage topographical information for precipitation interpolation?
%\item Does the difference in the interpolation results of input precipitation matter for climate impact modeling?
%\end{itemize}
%

To develop our network architecture we used SuperSloMo~\cite{Jiang+:CVPR18:SuperSloMo} (SSM), a multi-frame interpolation neural network, as our base network and applied the following adaptations:
\begin{itemize}
	\item We used self-supervised learning, specific to the spatio-temporal precipitation domain, while the original paper focused on general videos of 240fps for training and 30fps for evaluation input frames. 
	\item We used 3D vectors with an additional intensity (precipitation) transformation dimension, instead of 2D
optical flow vectors.
	\item We added additional topographical feature integration channels, including dot products to count ``rainy clouds climbing a hill''.
\end{itemize}

For unsupervised multiple video frames interpolation, we consider SSM as the state-of-the-art optical-flow approach whereas EpicFlow~\cite{Revaud+:CVPR15:EpicFlow} is one of the ``non-deep'' state-of-the-art approaches.
Deep Voxel Flow~\cite{Liu+:ICCV17:DeepVoxelFlow}, an early UNet-based (thus ``deep'') approach, demonstrated its advantage over EpicFlow.  SSM is an advanced version of Deep Voxel Flow, demonstrating equal-or-better performance over it with additional treatment of occlusions, etc.
%and thus can be regarded as one of the best unsupervised %multiple video frames interpolation approaches.

%The rest of this paper starts from motivating the problem setting with precipitation observation interpolation from satellite radar, for climate impact modeling. Then we describe the proposed interpolation method. The section of experimental results follows. After discussing related work, the paper finally concludes the remarks.
The rest of this paper first describes the proposed precipitation observation interpolation method along with the base one. Then, experimental results are presented, followed by some concluding remarks.

% remaining sections here after

%\input{motivation.tex}
\input{opticalflow.tex}
\input{experiment.tex}
%\input{relatedwork.tex}
% \input{demo.tex} % merged into experiment.tex

\section{Concluding Remarks}

We have introduced a novel adaptation of a multi-frame video interpolation technique for temporal interpolation of precipitation observation from satellite radars.   Terrain information is combined for use with temporarily coarse-grained precipitation radar observation as inputs for self-supervised training.  Our experiment with the  Meteonet dataset as ground-truth showed improved interpolation accuracy of our method, when compared to traditional linear interpolation.
% while the application to various select flooding events has shown the difference of precipitation interpolation could matter in calibrating flooding impact models, as we observe the difference in simulated flooding.
%We also observed the difference of precipitation interpolation matter in impact modeling.

% Future work
%Exploration of neural ordinary differential equations (ODEs) for optical flows and implicit neural representations for geometric features, instead of U-Net~\cite{Ronneberger+:MICCAI2015:U-Net} and topographic gradients respectively, would be an interesting future research venue.
%For example, application of neural ODEs to video interpolation for generating video frames at any given time steps has demonstrated high-quality inferred video frames in the continuous-time domain using real-world video datasets for both video interpolation and extrapolation.~\cite{Park+:AAAI2021:Vid-ODE}.
%Also, periodic activation functions are ideally suited for representing complex natural signals and their derivatives using implicit neural representations.~\cite{Sitzmann:NeurIPS2020:INR}

%\clearpage
%\scriptsize
% we customized bst for abbreviation
% https://tex.stackexchange.com/questions/430712/how-to-not-use-the-full-fist-name-in-ieee-spconf-paper-abbreviated-first-name-w
\bibliographystyle{IEEEbib-abbrev}
%\bibliography{bibs/interpolation,bibs/nowcasting,bibs/topography,bibs/modeling,bibs/opticalflow}

\input{main.bbl}
\end{document}

%% file: opticalflow.tex
%
% opticalflow.tex
% Methodology
\section{Temporal Precipitation Interpolation}

In this paper we leverage an approach introduced by SSM~\cite{Jiang+:CVPR18:SuperSloMo}, a high-quality variable-length
multi-frame interpolation method that can interpolate a video frame at any arbitrary time step between two frames, amongst many existing works on optical flow~\cite{Tu+:2019:CNN_optflow,Zhai+:2021:opticalflow_survey}.  While the simple application of SuperSloMo to satellite imagery has been recently studied~\cite{Vandal+Nemani:2021:SSM_satellite}, we extend it for the purpose of precipitation interpolation and the realistic availability of supplemental data of geography.

The main idea of the original SSM work is to warp the two input images to an arbitrary time step and then adaptively fuse the warped images to generate the intermediate image, where the motion interpretation and occlusion reasoning are modeled in a single end-to-end trainable network. Specifically, they first use a flow computation Convolutional Neural Network (CNN) to estimate the bi-directional optical flow between the two input images, which is then linearly fused to approximate the required intermediate optical flow in order to warp input images. To avoid poor approximations around motion boundaries, it uses another flow interpolation CNN to refine the flow approximations and predict soft visibility maps~\cite{Jiang+:CVPR18:SuperSloMo}.

While the experimental results in the SSM paper~\cite{Jiang+:CVPR18:SuperSloMo} use finer-grained training data (240fps) than prediction time (30fps), the training could have been done with the coarse-grained by considering intermediate frames in the consecutive frames or a clip as the interpolation target frames.  Also, since none of the learned network parameters are time-dependent, it can produce as many intermediate frames as needed.

\subsection{Base Intermediate Frame Synthesis}

Given two precipitation observation inputs $I_T \in \mathbb{R}^{H\times W}$ at times $T=0$ and $T=1$, and an intermediate time $t \in (0, 1)$, our goal is to predict the intermediate precipitation $\hat{I}_t  \in \mathbb{R}^{H\times W}$, at time $T=t$.
With multi-frames video interpolation approaches allowing arbitrary time  $t \in (0, 1)$ to interpolate at, such as SuperSloMo, the goal is to explicitly infer optical flows $F_{0 \rightarrow t}$ and $F_{1\rightarrow t}$, which are from $I_0$ to $I_t$, and from $I_1$ to $I_t$, respectively.

The goal is then to infer an inferred intermediate frame derived from a linear combination of warped $I_0$ and $I_1$ as:
\begin{equation}
\hat{I}_t = \alpha \odot g(I_0, F_{0 \rightarrow t})
			+ (1 - \alpha)  \odot g(I_1, F_{1\rightarrow t})
,
\end{equation}
where $g$ is a warping function.  The parameter $\alpha$ is a matrix of scalar weights $\in [0, 1]$ for taking the relative importance of the reference frames into account.  $\odot$ represents element-wise multiplication.

The backward warping function $g$ implemented using bilinear interpolation in SuperSloMo is defined as:
%
%  p.6 in https://arxiv.org/pdf/1605.03557.pdf
%\begin{equation}
%\begin{split}
%g^{(i)}(I, F) =  \\
%\sum_{ q \in
%	\text{neighbors of} (x'^{(i)}, y'^{(i)})
%}
%I^{(q)}
%(1 - |x'^{(i)} - x'^{(q)}|)
%(1 - |y'^{(i)} - y'^{(q)}|)
%\end{split}
%\end{equation}
\begin{equation} \label{eq:backwarp}
%\begin{split}
g^{(i)}(I, F) =  %\\
	\sum_{ q \in
		\text{neighbors of} (x'^{(i)}, y'^{(i)})
	}
						I^{(q)} w^{(q)}
%\end{split}
,
\end{equation}
\noindent
where $w^{(q)}$ is a weight, generally used in bilinear interpolation, for reflecting the proximity of a reference grid $q$ from the sampling location $(x'^{(i)}, y'^{(i)})$ for constructing a grid $i$.
It is defined using an optical flow $F^{(i)} := (\Delta x^{(i)}, \Delta y^{(i)}) $ and the target grid location $(x^{(i)}, y^{(i)})$ as $(x'^{(i)}, y'^{(i)}) =  (x^{(i)} - \Delta x^{(i)}, y^{(i)} - \Delta y^{(i)})$.
%:
%%
%\begin{equation}
%(x'^{(i)}, y'^{(i)}) =  (x^{(i)} - \Delta x^{(i)}, y^{(i)} - \Delta y^{(i)}) 
%.
%\end{equation}
%%

%TBA
%
%\begin{equation}
%\begin{split}
%F_{t\rightarrow 0} = \hat{F}_{t\rightarrow 0} 
%+ \Delta F_{t\rightarrow 0} \\
%F_{t\rightarrow 1} = \hat{F}_{t\rightarrow 1}
%+  \Delta F_{t\rightarrow 1}
%\end{split}
%\end{equation}
%
%\begin{equation}
%\begin{split}
%\hat{F}_{t\rightarrow 0}  = -(1 - t) t F_{0 \rightarrow 1} + t^2 F_{1 \rightarrow 0}  \\
%\hat{F}_{t\rightarrow 1} = (1 - t)^2 F_{0 \rightarrow 1} - t (1 - t) F_{1 \rightarrow 0} 
%\end{split}
%\end{equation}

\subsection{Intensity Change Consistency}

%~\cite{Horn+Schunck:1981:opticalflow}

%~\cite{Heas+:2008:3D-motion}

%If the level of brightness is not constant, then the motion estimate can be biased. To cope with this limitation, methods like the integrated continuity equation (ICE) have been developed and applied to satellite imagery (Fitzpatrick 1988; Corpetti et al. 2002; Heas et al. 2007). Other attempts have also been made to develop methods including brightness variation caused by time-dependent physical models (Haussecker and Fleet 2001), although these were mainly limited to relatively simple applications, such as changing illumination or thermal diffusion in infrared images.

In applications of unsupervised deep optical flow learning, originally found in ~\cite{Ren+:AAAI2017:unsupervised_opticalflow}, the reconstruction and photometric errors between the warped feature map from the reference image and the target image is treated as a loss to be back-propagated.  If the level of intensity is not constant, then the motion estimate can be biased~\cite{Tu+:2019:CNN_optflow}.

We introduce an additional dimension in the optical-flow vectors which represent intensity change.  Instead of $F = (\Delta x, \Delta y)$, now we rely on $F = (\Delta x, \Delta y, \Delta z)$, where $\Delta z$ represents the precipitation intensity increase from original grids.
The local consistency term, to be discussed in the following section, is applied to this additional dimension, as well as the original flow dimensions.

The backward warping function $g$, originally found in Equation~\ref{eq:backwarp}, is modified as:
\begin{equation}
%\begin{split}
g^{(i)}(I, F) =  %\\
\sum_{ q \in
	\text{neighbors of} (x'^{(i)}, y'^{(i)})
}
 (I^{(q)} + \Delta z^{(q)}) w^{(q)}
.
%\end{split}
\end{equation}

\subsection{Topographic Feature Integration}

\begin{figure*}[hbt]
	\centering
	\includegraphics[width=0.7\textwidth]{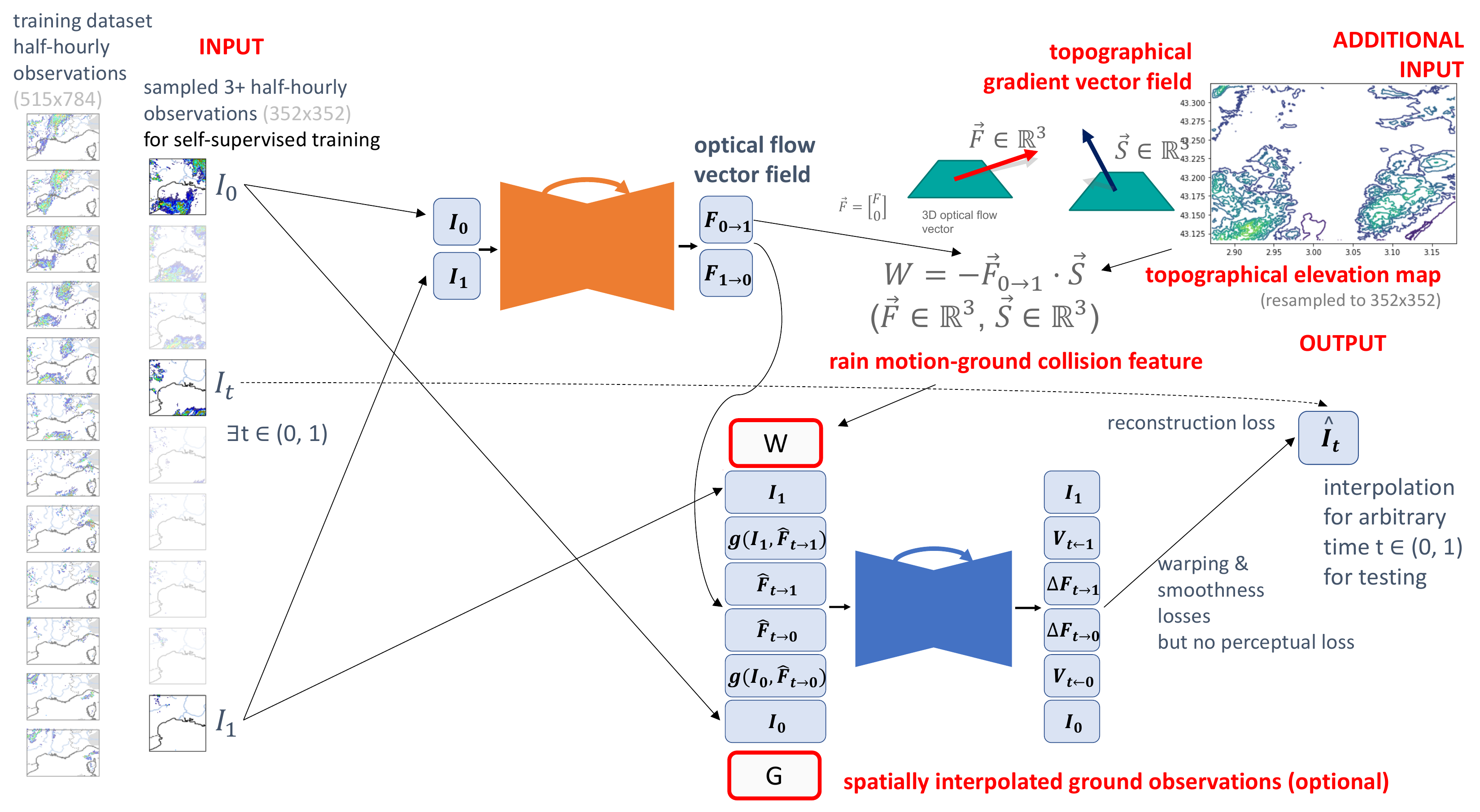}
	\caption{The proposed architecture for temporal precipitation interpolation based on a multi-frame video interpolation method~\cite{Jiang+:CVPR18:SuperSloMo} and an additional geospatial feature incorporated.}
	\label{fig:architecture}
\end{figure*}

In addition to the precipitation observation, we utilize the topographical elevation map to predict the interpolation, as depicted in Figure~\ref{fig:architecture}.  The entire network contains two U-Nets, which are fully convolutional neural networks. The first network takes two input images $I_0$ and $I_1$, to jointly predict the forward optical flow $F_{0 \rightarrow 1}$ and backward optical flow $F_{1 \rightarrow 0}$ between them. The second network then takes warped results at the target interpolation time $t$ in addition to the input images, to generate final optical flows and occlusion maps.  We extend the input to the second network with topographical elevation map.   In addition to feeding the map as is, we also include the feature engineered to represent the collision of rain cloud and the ground, by a dot-product of flow vector and terrain gradient.

\subsection{Training}

Instead of providing supervision for temporally fine-grained precipitation observation, which is generally difficult in practice, we train our network using the consequent observations in the original, low frequent observation of precipitation, say, consequent half-hourly observations.  While SuperSloMo has been tested with supervision, the design of network itself inherently has such a capability of self-supervision.

Given input precipitation frames $I_0$ and $I_1$, an intermediate
frame $I_t$ between them, where $t_i \in (0, 1)$, and the
prediction of intermediate frames $\hat{I}_t$, the loss function in SuperSloMo is a linear combination of four terms:
\begin{equation}
l = \lambda_r l_r + \lambda_p l_p + \lambda_w l_w + \lambda_s l_s
\end{equation}
\noindent
where$\lambda_r$, $\lambda_p$, $\lambda_w$ and $\lambda_s$ are weights for each loss term.  They are samely set as SuperSloMo.
%The weights have been set empirically using a validation set as λr =0.8,λp =0.005,λw =0.4, and λs =1. Every com- ponent of our network is differentiable, including warping and flow approximation. Thus our model can be end-to-end trained.

The reconstruction loss $l_r$ and the warping loss $l_w$ are defined in the same manner as SuperSloMo.  A reconstruction loss models the distance between $I_t$ and $\hat{I}_t$, whereas a warping loss models the quality of the computed optical flow.  The perceptual loss $l_p$,  preserves details of the predictions, however it is not used in this work as we do not have a pretrained model for precipitation. For comparison, SuperSloMo used the VGG16 model for general images.

The smoothness loss $l_s$, is used to encourage  neighboring pixels to have similar flow values as was done in SuperSloMo and is defined as :
\begin{equation}
l_s = ||\Delta F_{0 \rightarrow 1}||_1 + ||\Delta F_{1 \rightarrow 0}||_1
\end{equation}
\noindent
but our flows $F_{0 \rightarrow 1}$ and $F_{1 \rightarrow 0}$ are three dimensional.  The third dimension for intensity change has been introduced as described before.

%% file: experiment.tex
\begin{figure}[hbt]
	\centering
	\includegraphics[width=1.0\columnwidth]{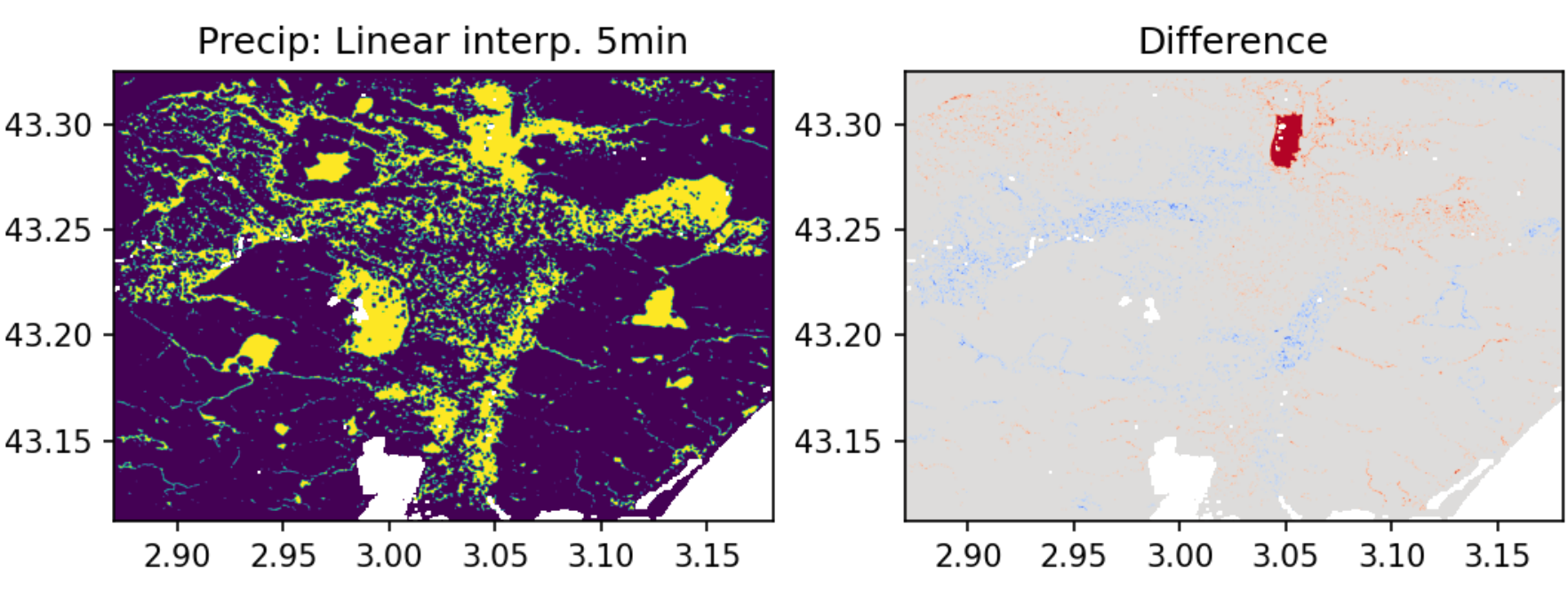}
	\caption{Flooding simulation results
		% at timestamp 2018-10-16 00:00:00
		using input precipitation
		%from the original fine temporal resolution (5 minutes interval, not shown) and
		from the linear interpolation (from 60 minutes to 5 minutes interval, left).
		%For illustration, a water depth threshold of 0.15\,m is used to define a pixel as flooded.
		The absolute difference from the result with the original fine temporal resolution (right) is mapped in colors (red - missing, blue - extra).}
	\label{fig:ifm_48h}
\end{figure}

\section{Experimental Results}

%\begin{figure*}[hbt]
%  \centering
%  \includegraphics[width=0.9\textwidth]{images/interpolation_1016.pdf}
%  \caption{Precipitation interpolation results at timestamp 2018-10-16 00:00:00. The groundtruth fine temporal resolution (5 minutes interval, top row), a linear interpolation result (middle), and the result of a proposed interpolation method (bottom).}
%  \label{fig:interpolation_1016}
%\end{figure*}

%\subsection{Dataset and Impact Modeling}

While the interpolation was designed to work with the globally available precipitation data from IMERG\footnote{\url{https://gpm.nasa.gov/data/imerg}}, the actual experiments and demonstration were conducted based on Meteonet\footnote{\url{https://meteonet.umr-cnrm.fr}} for validating the proposed interpolation with the ground truth precipitation observation, which is not generally available globally.  Samples extracted for every 30 minutes from the interpolation target month are used for training.  Every tuple of consequent 3 sampled frames in one hour (thus the middle frame as the self-supervising target) is given to a network in a random order.  Otherwise we follow the experimental setting of the original SSM\cite{Jiang+:CVPR18:SuperSloMo} but the training iteration was stopped at 20 epochs, where we observed general saturation of performance.

%10km / 0.1 Degree
%30min

To demonstrate the impact of various precipitation interpolation methods in flood simulation with IFM, we selected an actual flooding event in the past.  We select a flooding event in Aude, France back in October, 2018, which is the only flash flooding event covered by the Meteonet dataset time and area.  ``Several months' worth of rain fell within a few hours overnight in Aude, leaving people stranded on rooftops'', according to a news article by Sky News. In this event, a number of areas reportedly experienced flash flooding driven by heavy rainfall in a short period.
This resulted in the need for high resolution spatio-temporal rainfall information to support analysis with hydrological models to better understand what occurred on that night.

With the hypothesis that the ground topography of the area should have effects on the dynamics of the precipitation, we utilize a digital elevation model (DEM) released under the Shuttle Radar Topography Mission (SRTM)~\cite{van-Zyl:2001:SRTM}.
STRM is an international project aiming to generate high resolution topographic data globally.  Its DEM with a one arc-second (about 30 meters) resolution is publicly available for free.
%A number of numerical models has been utilizing the SRTM DEM since it was released. 
We consider the elevation as a relevant variable and incorporate it as the additional input layer channels of our interpolation refinement model.

\subsection{Preliminary Experiments for Impact Sensitivity}

Before looking at interpolation errors, we first confirm the sensitivity of flooding impact for precipitation input errors.  As a physical hydrological modeling system, we use the Integrated Flood Model (IFM), which offers fine-grained overland flood modeling with high scalability~\cite{Singhal+:EuroPar2014:IFM}.
%The architecture of IFM is illustrated schematically in Figure \ref{fig:ifm-system}. 
Since precipitation is an input driver for hydrological models such as IFM, the modeled results are directly affected by variations in the precipitation data.
%Therefore, we evaluate the performance of interpolation methods in terms of the simulation results driven by the resultant precipitation time series.

%Figure~\ref{fig:ifm_48h} shows the simulated flood extents at a single time stamp 48\,hours after the start of the simulation, obtained using precipitation data from both raw 5\,minutes resolution and 60-to-5\,minutes linearly interpolated resolution. The difference between these two binary maps, as shown on the right, reveals the difference in flood distribution, demonstrating the sensitivity of the flood model to the precipitation input data.
Figure~\ref{fig:ifm_48h} shows the simulated flood extents at a single time stamp 48\,hours after the start of the simulation, obtained using precipitation data from both raw 5\,minutes resolution and 60-to-5\,minutes linearly interpolated resolution. The difference between these two binary maps, as shown on the right, reveals the difference in flood distribution, demonstrating the sensitivity of the flood model to the precipitation input data.
%
%\begin{figure}[hbt]
%	\centering
%	\includegraphics[width=0.45\textwidth]{images/rmse_precip_flood.pdf}
%	%  \caption{Deviations in precipitation between the original 5 minutes resolution and the 60-to-5 minutes linearly interpolated resolution are shown by calculating the normalised root mean square error (RMSE) across the whole region of interest at each time step. The resulting impact on flooding is shown likewise by calculating the RMSE between the respectively simulated water depth maps. Note that the RMSE for flood is missing in the beginning because of too small water depths.}
%	\caption{Deviations in precipitation between the original 5 minutes resolution and the 60-to-5 minutes linearly interpolated resolution.}
%	\label{fig:rmse_precip_flood}
%\end{figure}
%
%In order to quantify the difference in the two precipitation data sets as well as in the resulting water depths, the root mean square errors (RMSE) are calculated across the whole area.
%Figure~\ref{fig:rmse_precip_flood} shows their normalised value over the whole time of the simulation period.It can be observed that the RMSE of the flooding results becomes maximum when the difference in precipitation becomes maximum. While the RMSE in precipitation is transient, the flooding depth is affected in the longer term, with a significant error remaining even after several days at the end of the simulation period. 
%
These results suggest that the difference of the interpolation methods applied to precipitation that drives hydrological models has potentially considerable impact to the flood simulation results, particularly in the case of flash floods that caused by local heavy rainfalls.

%We applied various temporal resolution of precipitation made from the MeteoNet dataset, 5-min resolution as is and half-hourly resolution by under-sampling.  While the interpolation has been applied in the overall area (South-East of France) of Meteonet, the evaluation is focused around Aude, a bounding rectangle where geographical coordinates of southwest and northeast corners in latitude and longitude are (43.1161, 2.8717) and (43.3218, 3.1795), respectively, so that major observed flash floods are included in the region of our analysis. For comparison, we also obtained simple linear interpolation results with the same setting as above. 
%The interpolated precipitation was also used to drive the IFM and flood depth was calculated as outputs of the simulation.

\subsection{Interpolation Results}

\begin{figure}[hbt]
	\centering
	\includegraphics[width=1.0\columnwidth]{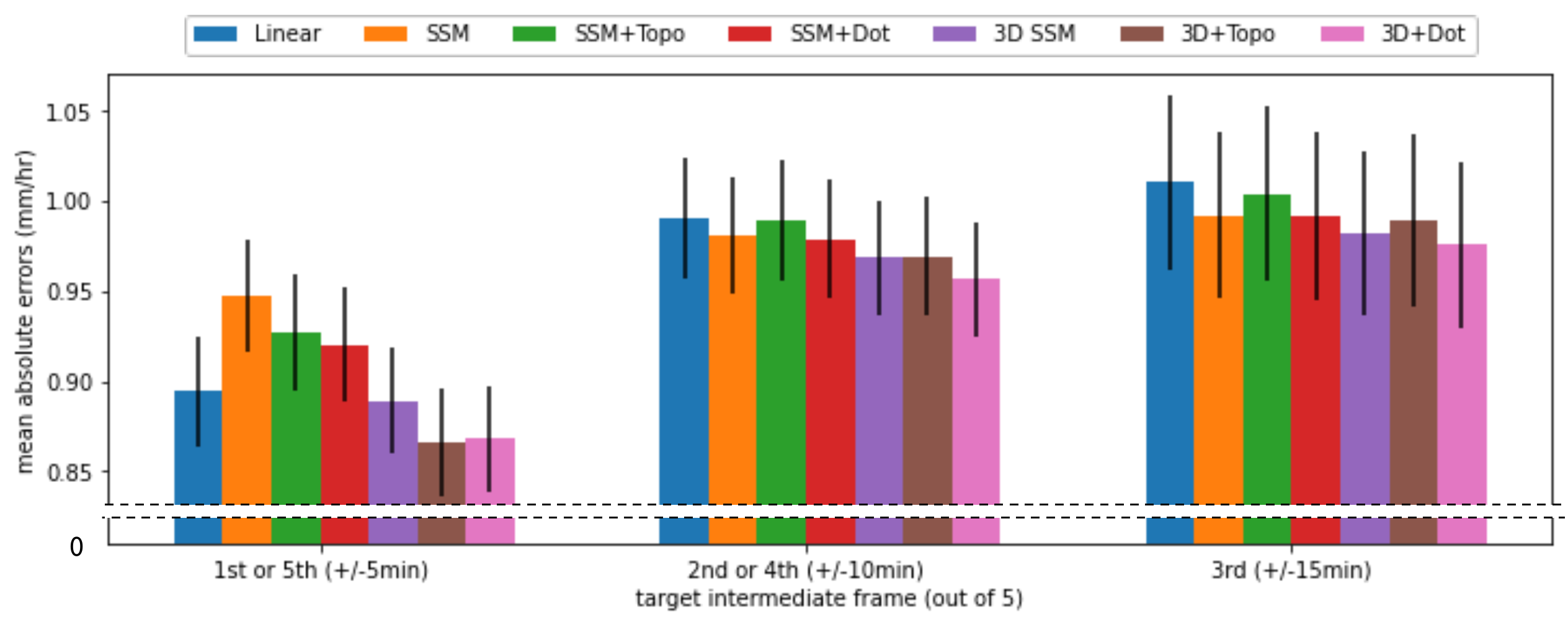}
	\includegraphics[width=1.0\columnwidth]{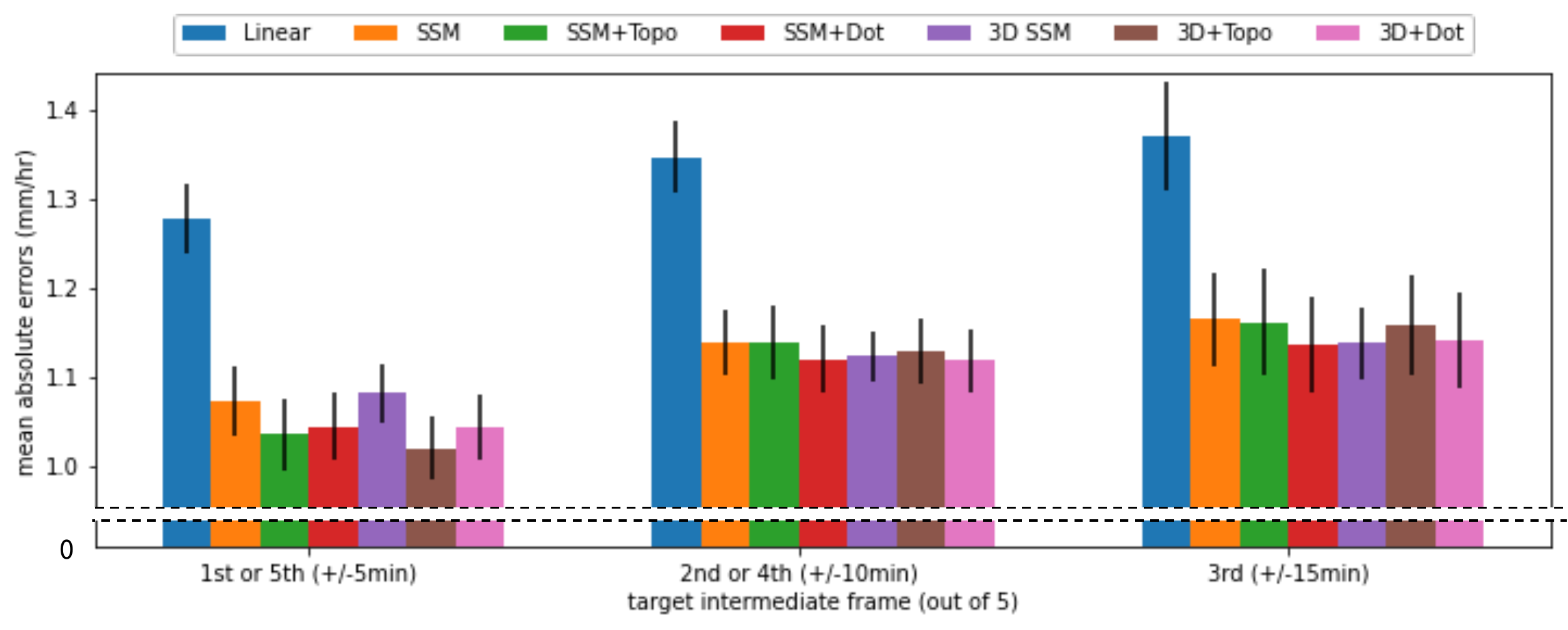}
	\caption{Mean absolute errors (lower better) of interpolated precipitation with SEM in average of 3 days (top, including the days before/after the flooding event) and 6 hours (bottom, just before the flooding event).}
	\label{fig:mae}
\end{figure}

Figure~\ref{fig:mae} shows the mean absolute errors from the ground truth precipitation, for each target interpolation time of 5 or 25, 10 or 20, and 15 minutes after the first reference frame.  Generally the baseline liner interpolation (Linear) is good at interpolating when the target is closer to reference frames.  Simply adapting SSM gives general reduction of errors.  Adding our proposed adaptation of topographic features (+Topo for additional DEM input as is, +Dot for dot products of flow and gradient vectors) and intensity change consistency (3D) gives further reduction.

The advantage of the proposed approach over the baseline linear interpolation (Linear) is the largest for interpolation temporarily far from the reference input frames, and for severe weather (6 hours before the flooding event), with up-to 20\% error reduction observed. The use of the intensity change (3D) generally matters for the 3-day average (including the day before and after the day of the flooding event), where  weather which isn't severe is included.  Supplying dot-product (of rain cloud flow and terrain gradient, Dot) features works reasonably well.

%\subsection{Historical Events}
%
%A total of 8 historical flooding events, from various locations across the globe were selected for the validation analysis (Table 1). All the flooding events where a result of extreme and often short-lived rainfall phenomenon, such as hurricanes, or other extreme weather systems. Table 1 provides a summary of the events, as well as the details of the model simulations. 
%
%\begin{table}
%	\caption{Events used in the validation}
%	
%\end{table}